# Title: Natural language processing to identify lupus nephritis phenotype in electronic health records

**Authors**: Yu Deng, BS[1], Jennifer A. Pacheco, MS[2], Anh Chung, BS,[1,6], Chengsheng Mao, PhD[1], Joshua C. Smith, PhD[3], Juan Zhao, PhD[3], Wei-Qi Wei, MD, PhD[3], April Barnado, MD, MSCI[4], Chunhua Weng, PhD[5], Cong Liu, PhD[5], Adam Cordon, PhD[2], Jingzhi Yu, BA[1], Yacob Tedla, PhD[1], Abel Kho, MD[1], Rosalind Ramsey-Goldman, MD[6], Theresa Walunas, PhD[1*], Yuan Luo, PhD[1*]

**Affiliations**: [1]Center for Health Information Partnerships, Feinberg School of Medicine, Northwestern University, Chicago, USA

[2]Center for Genetic Medicine, Feinberg School of Medicine, Northwestern University, Chicago, USA

[3]Department of Biomedical Informatics, Vanderbilt University Medical Center, Nashville, USA

[4]Department of Medicine, Vanderbilt University Medical Center, Nashville, USA

[5]Department of Biomedical Informatics, Columbia University, New York, USA

[6] Department of Medicine/Rheumatology, Feinberg School of Medicine, Northwestern University, Chicago, USA

* Dr. Luo and Dr. Walunas contributed equally to this work.

**Corresponding authors**: Yuan Luo, PhD, Theresa Walunas, PhD

**Email addresses:**

Yu Deng, yudeng2015@u.northwestern.edu, Jennifer A. Pacheco, japacheco@northwestern.edu, Anh Chung, a-chung@northwestern.edu, Chengsheng Mao, chengsheng.mao@northwestern.edu, Joshua C. Smith, joshua.smith@vumc.org, Juan Zhao, juan.zhao@vumc.org, Wei-Qi Wei, wei-qi.wei@uvmc.org, April Barnado, April.barnado@vumc.org, Chunhua Weng, cw2384@cumc.columbia.edu, Cong Liu, cl3720@cumc.columbia.edu, Adam Cordon, adam.gordon@northwestern.edu, Jingzhi Yu, k.yu@northwestern.edu, Yacob Tedla, yacob.tedla@northwestern.edu, Abel Kho, Abel.Kho@nm.org, Rosalind

Ramsey-Goldman, rgramsey@northwestern.edu, Theresa Walunas, t-walunas@northwestern.edu, Yuan Luo, yuan.luo@northwestern.edu

**Abstract**

Systemic lupus erythematosus (SLE) is a rare autoimmune disorder characterized by an unpredictable course of flares and remission with diverse manifestations. Lupus nephritis, one of the major disease manifestations of SLE for organ damage and mortality, is a key component of lupus classification criteria. Accurately identifying lupus nephritis in electronic health records (EHRs) would therefore benefit large cohort observational studies and clinical trials where characterization of the patient population is critical for recruitment, study design, and analysis. Lupus nephritis can be recognized through procedure codes and structured data, such as laboratory tests. However, other critical information documenting lupus nephritis, such as histologic reports from kidney biopsies and prior medical history narratives, require sophisticated text processing to mine information from pathology reports and clinical notes. In this study, we developed algorithms to identify lupus nephritis with and without natural language processing (NLP) using EHR data from the Northwestern Medicine Enterprise Data Warehouse (NMEDW). We developed four algorithms: a rule-based algorithm using only structured data (baseline algorithm) and three algorithms using different NLP models. The three NLP models are based on regularized logistic regression and use different sets of features including positive mention of concept unique identifiers (CUIs), number of appearances of CUIs, and a mixture of three components (i.e. a curated list of CUIs, regular expression concepts, structured data) respectively. The baseline algorithm and the best performed NLP algorithm were external validated on a dataset from Vanderbilt University Medical Center (VUMC). Our best performing NLP model incorporating features from both structured data, regular expression concepts, and mapped concept unique identifiers (CUIs) improved F measure in both the NMEDW (0.41 vs 0.79) and VUMC (0.62 vs 0.96) datasets compared to the baseline lupus nephritis algorithm.

**Introduction**

Systemic Lupus Erythematosus (SLE) is an autoimmune disease that has diverse manifestations, resulting in significant morbidity and mortality [1, 2]. While many autoimmune diseases, such as rheumatoid arthritis, have benefitted from new classes of medications, SLE has seen few advancements in therapy in the last 50 years [3]. It has been hypothesized that the heterogeneity of SLE presentations may make it challenging to understand therapeutic responses across the full scope of SLE presentations and that observational cohort studies and clinical trials would benefit from targeting subpopulations with similar disease presentations [4]. Recently, the Food and Drug Administration has approved two new medications for use in managing lupus nephritis, increasing the urgency of identifying lupus nephritis in people with SLE to ensure the new therapeutics can be targeted to these patients to help reduce kidney damage and improve long term outcomes [5]. Classification criteria for SLE describe a broad range of evidence-based clinical and laboratory descriptors. There are three criteria currently in use: 1) the set developed in 1982 and revised in 1997 by the American College of Rheumatology (ACR) [6], 2) the set developed by the System Lupus International Collaborating Clinics in 2012 (SLICC) [2], and 3) the set developed by the European League Against Rheumatism / American College of Rheumatology (EULAR/ACR) criteria set [7]. ACR classification criteria include malar rash, discoid rash, photosensitivity rash, oral ulcers, arthritis, serositis, glomerulonephritis, seizures, psychosis, Hematology, immunology, and anti-nuclear antibody (ANA). Patient is classified as SLE based on ACR if she/he meets at least 3 of the above-mentioned criteria. SLICC classification criteria include 11 clinical criteria (acute cutaneous lupus, chronic cutaneous lupus, oral ulcers, alopecia, synovitis, renal disorder, neurological disorder, hemolytic anemia, leukopenia, thrombocytopenia) and 6 immunological criteria (anti-nuclear antibody, anti-dsDNA, anti-smith, antiphosphoilipid antibody, low complement, direct Coombs test). Lupus nephritis is one of the most common and severe manifestations of SLE where approximately 40% of SLE patients develop lupus nephritis [8], and it is included in all three classification criteria sets. In both the SLICC and EULAR/ACR criteria, one way to be classified as "definite lupus" is having a positive anti-nuclear antibody/anti-dsDNA screen in the presence of renal biopsy-proven lupus nephritis [2, 7]. Thus, lupus nephritis is a critical attribute to describe for clinical and research applications and the development of SLE subpopulations, but often it requires time consuming chart adjudication to identify patients who satisfy this criterion.

Electronic health records (EHRs) are a readily available data source that includes a record of clinical care and procedures, diagnoses, laboratory test results, medication orders, and clinical notes for describing disease manifestations in persons with SLE. However, a large amount of information in the EHR, such as histology notes for kidney biopsies, is generally only located in text-based notes from which it is challenging to extract information using simple rule-based identification algorithms and text string searches [9, 10]. Several prior studies developed algorithms to identify lupus nephritis using administrative claims data [11]. Chibnik et al. identified lupus nephritis in claims data and reached a positive predictive value (PPV) of 88% but sensitivity and specificity were not mentioned [12]. Li et al. used various combinations of International Classification of Diseases (ICD) codes to identify lupus nephritis [13]. Their algorithm achieved good sensitivity and specificity but a low positive predictive value (PPV) of 63.4%. Most of these studies only used structured data (i.e. ICD codes, laboratory test value), and the algorithms were often not validated in an external dataset [12, 13]. Thus, correctly identifying lupus nephritis from EHR for large cohort studies, in addition to identifying critical procedures, diagnoses and lab results, also requires the development of natural language processing (NLP) tools that can utilize histology reports and clinical notes. Previously, studies with other structured data-based concepts (e.g. multiple sclerosis, rheumatoid arthritis) have demonstrated that NLP can significantly improve rate of identification [9, 14].

In this study, we compared algorithms for the identification of lupus nephritis in SLICC criteria based on structured data alone with those that included three different NLP models to determine whether NLP could improve identification of persons with lupus nephritis. Our approach facilitates accurate identification of lupus nephritis in the EHR enabling researchers to better understand patients' SLE characteristics and serving as a foundation for lupus nephritis-related large cohort observational studies and clinical trials. We trained and evaluated the performance of all four algorithms in a dataset from Northwestern Medicine Electronic Data Warehouse (NMEDW) and then further validated the performance in an external dataset from Vanderbilt University Medical Center (VUMC).

## Methods

### Data source

The Chicago Lupus Database (CLD), established in 1991, is a physician validated registry of 1,052 patients with possible or definite lupus according to the 1982 American College of Rheumatology classification criteria revised in 1997 [15, 16]. The patients in the CLD met at least three ACR criteria (step 1 in Figure 1). Among the 1052 patients in the CLD, 878 patients had definite lupus according to the Systemic Lupus International Collaborating Clinics (SLICC) classification criteria (step 2 in Figure 1) [2]. Among these patients, 178 have lupus nephritis according to the definition in SLICC. The presence or absence of lupus nephritis in patients in the CLD is verified by the physician chart review.

The Northwestern Medicine Electronic Data Warehouse (NMEDW) is the primary data repository for all the medical records of patients who receive care within the Northwestern Medicine system [17]. Established in 2007, the NMEDW contains records for over 3.8 million patients, with most EHR data going back to at least 2002, and with some billing claims data going back to 1998 or earlier. By linking patients in the CLD to patient records in the NMEDW through their medical record numbers, we identified 818 definite SLE patients based on SLICC criteria who were both in the CLD and the NMEDW (see step 3 in Figure 1). To ensure our patient cohort has sufficient depth of data in both data sources, we excluded any patients who had less than four clinical encounters documented in the NMEDW [18, 19], reducing the final case cohort size to 472 (see step 4 in Figure 1). All inpatient and outpatient notes from transplant, nephrology, and rheumatology departments were retrieved. The retrieved clinical narratives included pathology reports, progress notes, consult notes, and discharge notes.

**Algorithm development**

Lupus nephritis is defined as "having a urine protein/creatinine ratio (or 24-hour urine protein collection) equivalent to 500 mg of protein per 24-hour period, or red blood cell casts in the urine" based on the SLICC classification criteria [2]. We developed four algorithms (see Table 1 for the overview of the four algorithms) to identify lupus nephritis from SLE patients' EHR data including a baseline algorithm that used only structured data and three NLP models that used structured data and clinical notes. In the baseline algorithm, a patient is classified as lupus nephritis based on ICD9/10 diagnosis codes and laboratory test results. The details of the structured data used in the baseline algorithm are shown in Supplemental Table 1. For the NLP models, we implemented an L2-regularized logistic regression classifier. We chose L2-regularized logistic regression because it can handle high dimensional feature space and multicollinearity problems by penalizing its coefficients in the loss function. In addition, the model is straightforward, and model output is easy to interpretate. We tried both L1 and L2-regularized logistic regression and selected the latter because it generates equivalent if not superior performance compared to L1-regularized logistic regression in our NU dataset. Following the steps in Zeng et al. [20-22], we extracted concept unique identifier (CUI) features and regular expression (regex) matches from the notes. For the CUI features, we first preprocessed the notes by removing duplicated records and tokenizing sentences. We then applied MetaMap to annotate medical concepts in each sentence [23]. MetaMap is an NLP application that maps biomedical text to the Unified Medical Language System (UMLS) Metathesaurus and assigns a CUI to each word or term [24]. Any CUIs recognized as being negated by MetaMap (i.e., "no glomerulonephritis") were excluded. For regex features, five concepts were used as features, including nephritis class II, nephritis class III, nephritis class IV, nephritis class V, and proteinuria. We developed regular expression patterns to search for text related to the five concepts (see Supplemental Table 2 for the list of regex patterns). For the NLP models, we developed three models using different feature sets. In the first NLP model– the full MetaMap (binary) model, all positive mentioned MetaMap CUIs were used as binary type features. In the second NLP model– the full MetaMap (count) model, the number of occurrences for every positive mapped CUIs were used as features. The minimum document frequency was set as 30 and 40 in MetaMap (binary) model and MetaMap (count) model, respectively to avoid feature sparsity. The frequencies were chosen by trying a list of frequencies and the ones generated the highest F measure were selected. In the third NLP model– the MetaMap mixed model, we used a mixture of lupus nephritis related CUIs, structured data, and regex concepts as features. The CUIs include C0024143, C0268757, C0268758, C4053955, C4053958, C4053959, C4054543 (see Supplemental Table 3 for each CUI definition). For the structured data component, a single binary feature is used. A patient is indicated to be positive for the structured data feature if he/she is predicted positive in the baseline algorithm. There were 13 variables in total for the MetaMap mixed model including 7 features from CUIs, 5 lupus nephritis related concepts for regex expression search, and 1 feature from structured data.

**Model training and evaluation**

We split the data from NMEDW into training (75%) and testing datasets (25%). In the training dataset, to get the optimal hyperparameter, we used grid search on parameter *C,* which is the inverse of regularization strength, ranging from 1e-5 to 1e5 with interval spacing equal to 10. We selected "sag" method as our optimizer [25]. We set the class weight as balanced to adjust for disproportionate class frequencies. Parameters that generated the best accuracy were retained. We evaluated our model in the testing set (internal validation) based on sensitivity, specificity, PPV, negative predictive value (NPV), F measure, and area under the curve (AUC). We further explored feature contribution by extracting the top 5 features with the highest positive coefficient in MetaMap (binary), MetaMap (count), and MetaMap mixed model, respectively. We also evaluated feature importance by generating mean absolute Shapley value (SHAP) plots. L2-regularized logistic regression was conducted using 'scikit-learn' library in Python, version 3.7.3. Regular expression was performed using 're' package in Python, version 3.7.3 [25, 26]. Shapley value was generated using 'shap' package in Python, version 3.7.3. SHAP plot was generated using 'matplotlib' package in Python, version 3.7.3.

**External validation**

We further validated both the baseline algorithm, and the best performing NLP model (based on results from the testing set at Northwestern University site) in an external validation dataset at Vanderbilt University Medical Center (VUMC), a regional, tertiary care center [27, 28]. The VUMC data warehouse contains over 3.2million subjects with de-identified clinical records from the EHR collected across the past several decades. We first performed a simple SLE phenotyping algorithm based on SLE ICD9/10 codes to get a SLE cohort (not chart reviewed) on which to run our lupus nephritis algorithm. We then randomly selected 50 patients on which to evaluate our lupus nephritis algorithm. A rheumatologist manually reviewed the chart for these 50 patients. Among these patients, there were 16 patients with definite lupus, 1 with possible SLE, and 33 with no SLE. There were 13 patients with lupus nephritis and 37 patients without. We evaluated the F measure, sensitivity, specificity, PPV, and NPV for the lupus nephritis baseline algorithm and NLP model with the highest F measure based on the results from the Northwestern University (NU) dataset. F measure evaluates the accuracy of the algorithm, it is calculated as the following:

$$F\ measure = \frac{2*precision*recall}{precision+recall},$$

Here precision and recall are also known as PPV and sensitivity, respectively.

## Results

Among the 472 SLE patients at NU, there were 178 patients (37.7% of the cohort) who developed lupus nephritis. The average number of notes per patient is 68.58 (standard deviation [SD] = 59.37). The distribution of the number of notes for the patient cohort is shown in Figure 2.

The performance for the four algorithms is shown in Table 2. All three NLP models have higher sensitivity, specificity, PPV, and NPV compared to the baseline algorithm using structured data alone. The full MetaMap (binary) model has higher sensitivity compared to the full MetaMap (count) model, (0.63 vs 0.6), NPV (0.81 vs 0.8), and comparable F measure (0.71 vs 0.71). The MetaMap mixed model has higher sensitivity (0.74) and NPV (0.86) as well as F measure (0.79) compared to the other two models. Similarly, MetaMap mixed model has higher AUC (0.89) compared to full MetaMap (binary) (AUC = 0.85) and full MetaMap (count) (AUC = 0.84) model (see Figure 3). Therefore, we selected the MetaMap mixed model as the final NLP model to be validated at VUMC in addition to the baseline algorithm. In the VUMC dataset, which included 50 patients, the MetaMap mixed model has higher sensitivity, specificity, PPV, and NPV compared to the baseline algorithm. The F measure improved from 0.79 to 0.96 as shown in Table 2.

In terms of feature importance, the top 5 features with the highest positive coefficient for each classifier are shown in Table 3. C0024143 (lupus nephritis) appears to have high positive coefficient in all three L2-regualrized classifiers. C1962972 (proteinuria finding) are the 4$^{th}$ highest positive coefficient in both MetaMap (binary) and MetaMap (count) model. Our full MetaMap models are able to pick up many important lupus nephritis related concepts such as kidney disease, proteinuria, lupus nephritis as high coefficient features. The SHAP plot shows the top 10 most important features for classification in each model. As shown in Figure 4, 5, 6, most of the important features are related to lupus nephritis clinically.

## Discussion

In this study, we developed four algorithms to identify lupus nephritis: a baseline algorithm using structured data only, a full MetaMap model with binary features, a full MetaMap model with count features, and a MetaMap mixed model. In the NU testing dataset, the MetaMap mixed model outperformed (F measure = 0.79) both the baseline algorithm (F measure = 0.41) and the other two NLP models (F measure = 0.60, 0.63 respectively). In the VUMC validation dataset, the MetaMap mixed model significantly improved the F measure over the baseline algorithm (0.96 versus 0.62).

*Error analysis*

In the MetaMap mixed model, we investigated 10 SLE patients in the training set that were wrongly classified by L2-regularized logistic regression. One patient was wrongly predicted as negative for lupus nephritis with a 0.49 probability of having lupus nephritis. In the feature set the algorithm identified, the patient was positive for CUI C002413 (glomerulonephritis in the context of systemic lupus erythematosus) and was negative for all the other features. It was mentioned in the notes that the patient had 'stage 2 LN'. Lupus nephritis class II is one of the features used in our algorithm. However, our regex did not include this specific variation of wording for lupus nephritis class II. This pattern could be incorporated in the NLP in the future to improve algorithm performance.

In another example, a 26-year-old female was wrongly predicted as positive for lupus nephritis with a probability of 0.53 of having lupus nephritis. In the feature set the algorithm identified, the patient was positive for C0024143 (glomerulonephritis in the context of systemic lupus erythematosus) and proteinuria features both of which were positively associated with lupus nephritis. Our algorithm showed that the patient had matched for 'proteinuria>0.5' in the notes which was in the context of 'negative renal disorder: either persistent proteinuria (>0.5g/day or +++) or cellular casts'. Our regex pattern was not able to capture the negation at the beginning of the sentence. Therefore, it falsely predicted the patient as positive for lupus nephritis.

All NLP models outperformed the baseline algorithm in the NU testing (internal validation) dataset. In the baseline model, 20/35 lupus nephritis patients were wrongly classified as non-lupus nephritis patients, while the MetaMap mixed model reduced the misclassified cases to 9/35. The baseline algorithm relies solely on ICD 9/10 diagnosis and laboratory test results. In the baseline rule-based algorithm, laboratory tests missing from the EHR largely influenced the performance. In the NLP MetaMap mixed model, using features from multiple modalities (EHR notes-derived regex, CUIs features, laboratory tests, and ICD codes) that complement each other, and a penalized logistic regression model improved the accuracy and generalizability of the model. As part of the future work, we plan to apply advanced imputation methods [29, 30] to fill in missing laboratory tests in order to further improve the phenotyping performance.

*Limitations*

Our study has certain limitations. We only had 50 patients in the VUMC validation dataset. This is due to limited resources for chart review. The small sample size may increase the chance of sample bias, which might explain the big improvement of F-measure in the external validation dataset. Future study is needed to further validate our algorithm performance in a larger external dataset.

## Conclusion

In conclusion, we developed four algorithms, a structured data only algorithm and three NLP models, to identify lupus nephritis phenotypes. We evaluated the algorithms in an internal and an external validation dataset. All three NLP models outperformed the baseline algorithm in the internal validation dataset. In the external validation dataset, our NLP MetaMap mixed model improved the F-measure greatly compared to the structured data only algorithm. Our NLP algorithms can serve as powerful tools to accurately identify lupus nephritis phenotype in EHR for clinical research and better targeted therapies.

## Abbreviations

**SLE:** Systemic lupus erythematosus

**EHRs:** electronic health records

**NLP:** natural language processing

**NMEDW:** Northwestern Medicine Enterprise Data Warehouse

**VUMC:** Vanderbilt University Medical Center

**CUI:** concept unique identifier

**ACR:** American College of Rheumatology

**SLICC:** System Lupus International Collaborating Clinics

**EULAR/ACR:** European League Against Rheumatism / American College of Rheumatology

**ICD**: International Classification of Disease

**PPV**: positive predictive value

**NPV**: negative predictive value

**CLD**: Chicago Lupus Database

**Regex**: regular expression

**UMLS**: Unified Medical Language System (UMLS) Metathesaurus

**NU**: Northwestern University

## Declarations

### *Ethics approval and consent to participate*

This study was a retrospective study of existing records. Ethics approval was provided by Northwestern University Institutional Review Board and Vanderbilt University Institutional Review Board.

### *Consent for publication*

Not applicable

### *Availability of data and materials*

The datasets generated and analyzed during the current study are not publicly available due to protected patient information but are available from the corresponding author on reasonable request.

### *Competing interests*

The authors declare that they have no competing interests.

### *Funding*

This work was supported by National Institute of Arthritis and Musculoskeletal and Skin Diseases (NIAMS 5R21AR072263), NIH/NIAMS R61 AR076824-01, NIH/NIAMS 1K08 AR072757-01 and Rheumatology Research Foundation K Supplement Award. This project is part of the Phase III of the eMERGE Network was initiated and funded by the NHGRI through the following grants: U01HG008672 (Vanderbilt University Medical Center); U01HG008680 (Columbia University Health Sciences); and U01HG008673 (Northwestern University). The publication fee will be charged through the eMERGE funding (U01HG008673). The funding body does not take part in the design of the study and collection, analysis, and interpretation of data and writing the manuscript.

### *Authors' contributions*

YD led the study, performed all data analyses, and wrote the manuscript. YL and TW designed the study. YL and TW supervised the project. RRG provided clinical expertise on interpreting the data. JAP coordinated the project, assisted with the design of the algorithms, implemented the algorithm without NLP, and provided expertise in EHR data analysis. RRG, TW, and AC made substantial contribution to the data acquisition. All the other authors read, edited, and approved the final manuscript.

*Acknowledgements*

Not applicable.

*Corresponding author*

Correspondence to Yuan Luo, Theresa Walunas.

**Figure 1.** SLE case cohort selection process. We identified 1052 SLE patients who met at least 3 ACR criteria based on physician chart review. Among these 1052 patients, we further identified 878 patients who also met SLICC classification criteria. Among the 878 patients, 818 patients were in NMEDW. We further restricted our study cohort to patients who had at least 4 encounters in the NMEDW which left 472 patients in the final cohort. Abbreviations: ACR criteria, American College of Rheumatology Classification Criteria; CLD, Chicago Lupus Database; SLE, systemic lupus erythematosus; NMEDW, Northwestern Medicine Enterprise Data Warehouse.

**Figure 2**. Histogram of note count per patient for SLE patients.

**Figure 3.** Area under the curve (AUC) for Full MetaMap (binary), Full MetaMap (counts), and MetaMap mixed model in NU testing set.

**Figure 4.** SHAP plot for full MetaMap (binary) model with SHAP feature importance measured as the mean absolute Shapley values.

**Figure 5.** SHAP plot for full MetaMap (count) model with SHAP feature importance measured as the mean absolute Shapley values.

**Figure 6.** SHAP plot for MetaMap mixed model with SHAP feature importance measured as the mean absolute Shapley values.

**Table 1.** Algorithm description

| Algorithm name | Classification model | Description |
|---|---|---|
| Baseline algorithm | Rule-based | A patient is confirmed to have lupus nephritis if he/she has proteinuria>0.5mg in laboratory test or has ICD 9/10 diagnosis code for lupus nephritis. |
| Full MetaMap model (binary) | L2-regularized logistic regression | Features are the non-negative mention of MetaMap CUIs. We treated CUIs as binary variables and fitted L2-regularized logistic regression to predict lupus nephritis. |
| Full MetaMap model (count) | L2-regularized logistic regression | The same as the full MetaMap model (binary) except that MetaMap CUIs are treated as numeric variables representing the count of instances each concept is mentioned in the clinical text. |
| MetaMap mixed model | L2-regularized logistic regression | There are 13 features in this model including 7 CUI features, 5 RegEx concepts, and 1 feature from structured data. |

**Table 2.** Model performance

| Phenotype | Dataset | Algorithm | Sensitivity | Specificity | PPV | NPV | F Measure |
|---|---|---|---|---|---|---|---|
| Lupus nephritis | NU (testing set) | Baseline | 0.43 | 0.6 | 0.39 | 0.64 | 0.41 |
| Lupus nephritis | NU (testing set) | Full MetaMap (binary) | 0.63 | 0.92 | 0.82 | 0.81 | 0.71 |
| Lupus nephritis | NU (testing set) | Full MetaMap (counts) | 0.6 | 0.95 | 0.88 | 0.8 | 0.71 |
| Lupus nephritis | NU (testing set) | MetaMap mixed | 0.74 | 0.92 | 0.84 | 0.86 | 0.79 |
| Lupus nephritis | VUMC | Baseline | 0.92 | 0.61 | 0.46 | 0.96 | 0.62 |
| Lupus nephritis | VUMC | MetaMap mixed | 1 | 0.97 | 0.93 | 1 | 0.96 |

**Abbreviations:** SLE, systemic lupus erythematosus; NU, Northwestern University; VUMC, Vanderbilt University Medical Center; NLP: natural language processing; PPV, positive predictive value; NPV, negative predicted value. For logistic regression-based models, probability of 0.5 is used as the threshold for classification.

**Table 3.** Top 5 positive coefficient for each classifier.

| Coefficient ranking | Full MetaMap (binary) | | Full MetaMap (count) | | MetaMap Mixed | |
|---|---|---|---|---|---|---|
| | Feature, definition | Coefficient | Feature, definition | Coefficient | Feature, definition | Coefficient |
| 1 | C0027697, nephritis | 0.04 | C0022646, kidney | 5.01 | C0024143, lupus nephritis | 1.26 |
| 2 | C0024143, lupus nephritis | 0.04 | C0024143, lupus nephritis | 4.79 | 'RENAL' | 0.64 |
| 3 | C0022658, kidney disease | 0.03 | C0033687, proteinuria | 3.58 | 'nephritis class IV' | 0.54 |
| 4 | C1962972, proteinuria finding | 0.03 | C1962972, proteinuria finding | 3.53 | 'proteinuria>0.5 gm' | 0.45 |
| 5 | C003368, sultroponium | 0.03 | C1707664, Delayed Release Dosage Form | 3.52 | C4053955, SLE class IV | 0.25 |

Features are ranked by the value of their associated coefficients. RENAL, renal indictor from structured data only; gm, gram.

**Additional files**

**File name:** Supplemental tables

**File format:** word document

**Titles and descriptions:**

Table 1. ICD-9/10 codes and LOINC codes used for baseline algorithm.

Table 2. Regex concepts and their associated searching keywords for lupus nephritis.

Table 3. CUIs and their definition